\pdfoutput=1

\documentclass[11pt]{article}

\usepackage{acl}

\usepackage{times}
\usepackage{latexsym}

\usepackage[T1]{fontenc}

\usepackage[utf8]{inputenc}

\usepackage{microtype}

\title{PHALM: Building a Knowledge Graph from Scratch\\by Prompting Humans and a Language Model}

\author{Tatsuya Ide$^1$ \And Eiki Murata$^1$ \And Daisuke Kawahara$^1$
        \AND
        Takato Yamazaki$^2$ \And Shengzhe Li$^2$ \And Kenta Shinzato$^2$ \And Toshinori Sato$^2$
        \AND \\
        $^1$Waseda University ~~$^2$LY Corporation\\
        \small \texttt{\{t-ide@toki.,eiki.murata.1650-2951@toki.,dkw@\}waseda.jp}\\
        \small \texttt{\{takato.yamazaki,shengzhe.li,kenta.shinzato,toshinori.sato\}@lycorp.co.jp}
        }

\usepackage{graphicx}
\graphicspath{ {./images/} }
\usepackage{caption}
\usepackage{subcaption}
\usepackage{CJKutf8}
\usepackage{amsmath}
\usepackage{amssymb}
\usepackage {arydshln}
\usepackage{hyperref}

\begin{document}
\maketitle
\begin{abstract}
Despite the remarkable progress in natural language understanding with pretrained Transformers, neural language models often do not handle commonsense knowledge well.
Toward commonsense-aware models, there have been attempts to obtain knowledge, ranging from automatic acquisition to crowdsourcing.
However, it is difficult to obtain a high-quality knowledge base at a low cost, especially from scratch.
In this paper, we propose PHALM, a method of building a knowledge graph from scratch, by prompting both crowdworkers and a large language model (LLM).
We used this method to build a Japanese event knowledge graph and trained Japanese commonsense generation models.
Experimental results revealed the acceptability of the built graph and inferences generated by the trained models. We also report the difference in prompting humans and an LLM.
Our code, data, and models are available at \hyperlink{https://github.com/nlp-waseda/comet-atomic-ja}{github.com/nlp-waseda/comet-atomic-ja}.
\end{abstract}

\section{Introduction}
\label{sec:intro}

Since pretrained models \citep{Radford2018ImprovingLU, devlin-etal-2019-bert, NEURIPS2019_dc6a7e65} based on Transformer \citep{NIPS2017_3f5ee243} appeared, natural language understanding has made remarkable progress.
In some benchmarks, the performance of natural language understanding models has already exceeded that of humans.
These models are applied to various downstream tasks ranging from translation and question answering to narrative understanding and dialogue response generation.
In recent years, the number of parameters in such models has continued to increase \citep{Radford2019LanguageMA, NEURIPS2020_1457c0d6}, and so has their performance.

When we understand or reason, we usually rely on commonsense knowledge.
Computers also need such knowledge to answer open-domain questions and to understand narratives and dialogues, for example.
However, pretrained models often do not handle commonsense knowledge well \citep{Zhou_Zhang_Cui_Huang_2020,Hwang_Bhagavatula_Le_Bras_Da_Sakaguchi_Bosselut_Choi_2021}.

There are many knowledge bases for commonsense inference.
Some are built by crowdsourcing \citep{Speer_Chin_Havasi_2017, Sap_Le_Bras_Allaway_Bhagavatula_Lourie_Rashkin_Roof_Smith_Choi_2019, Hwang_Bhagavatula_Le_Bras_Da_Sakaguchi_Bosselut_Choi_2021}, but acquiring a large-scale knowledge base is high-cost.
Others are built by automatic acquisition \citep{https://doi.org/10.48550/arxiv.1905.00270, ijcai2020p554}, but it is difficult to acquire high-quality commonsense knowledge.
Recently, there have been some methods using large language models (LLMs) for building knowledge bases \citep{NEURIPS_DATASETS_AND_BENCHMARKS2021_9fc3d715, west-etal-2022-symbolic, https://doi.org/10.48550/arxiv.2201.05955}.  They often extend existing datasets, but do not build new datasets from scratch. Furthermore, there are not many non-English datasets, although the knowledge that is considered commonsense varies by languages and cultures \cite{lin-etal-2021-common,Nguyen-cultural,Acharya2020AnAO}.

\begin{figure}[t]
    \centering
    \includegraphics[width=0.9\linewidth]{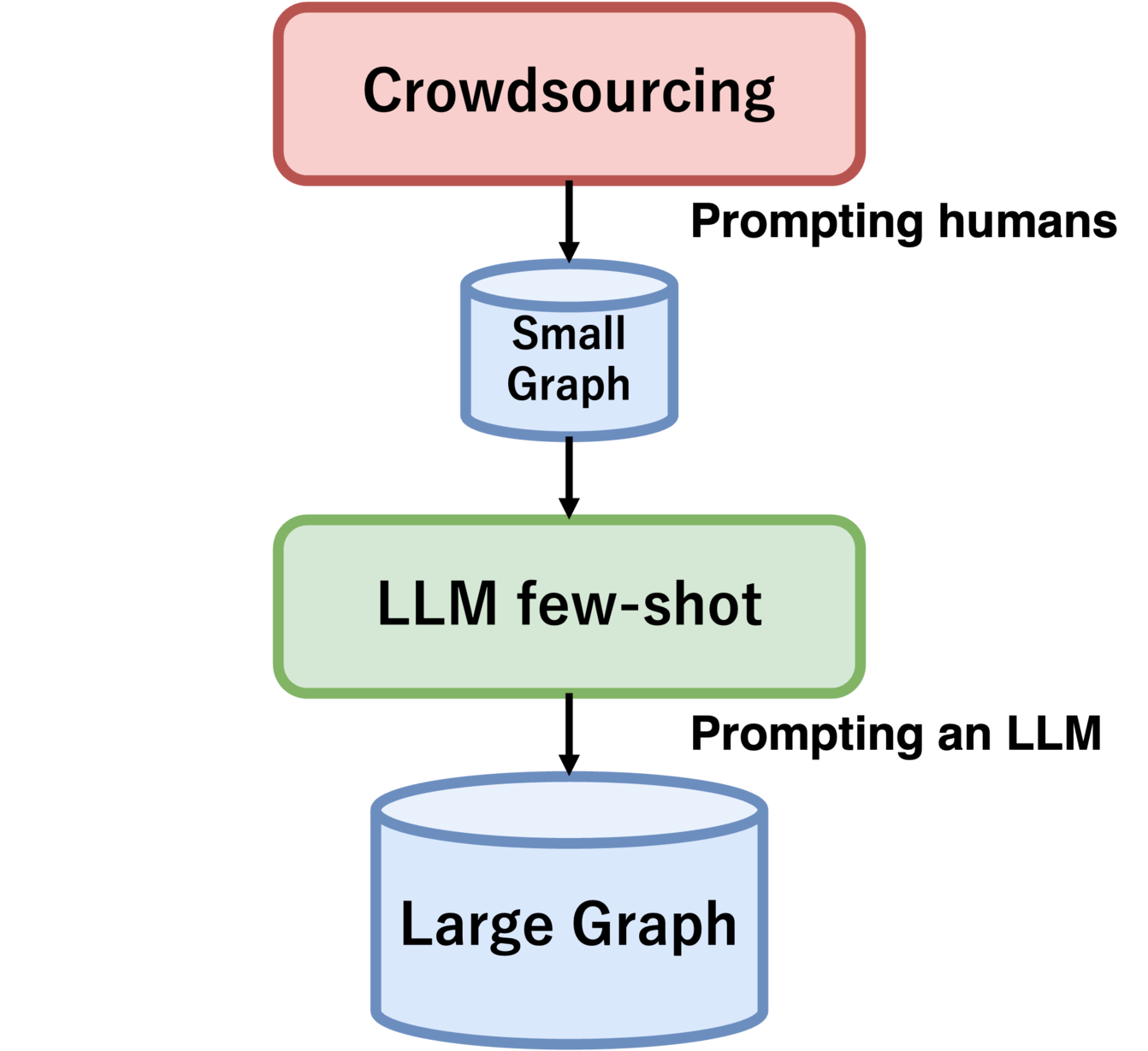}
    \caption{An overview of our method. We build a knowledge graph step by step \textbf{from scratch}, by prompting both humans and an LLM.}
    \label{fig:method}
\end{figure}

In this paper, we propose PHALM\footnote{\textbf{P}rompting \textbf{H}umans \textbf{A}nd a \textbf{L}anguage \textbf{M}odel.}, a method to build a knowledge graph from scratch with both crowdsourcing and an LLM.
Asking humans to describe knowledge using crowdsourcing and generating knowledge using an LLM are essentially the same (as it were, the latter is an analogy of the former), and both can be considered to be \textit{prompting}.
Therefore, we consider prompting for both humans and an LLM and gradually acquire a knowledge graph from a small scale to a large scale.
Specifically, we acquire a small-scale knowledge graph by asking crowdworkers to describe knowledge and use them as a few shots for an LLM to generate a large-scale knowledge graph.
At each phase, we guarantee the quality of graphs by applying appropriate filtering. In particular, we also propose a low-cost filtering method for the commonsense inferences generated by the LLM.

We built a Japanese knowledge graph on events, considering prompts for both humans and an LLM.
With Yahoo! Crowdsourcing\footnote{\url{https://crowdsourcing.yahoo.co.jp/}} and HyperCLOVA JP, a Japanese variant of the LLMs built by \citet{kim-etal-2021-changes}, we obtained a knowledge graph that is not a simple translation, but unique to the culture.
Then, we compared inferences collected by crowdsourcing and generated by the LLM. In addition to acquisition, we trained a Japanese commonsense generation model based on the built knowledge graph.
With the model, we verified the acceptability of output inferences for unseen events.
The resulting knowledge graph and the commonsense model created in this paper will be released to the public.

In summary, our contributions are (1) the proposal of a method to build a large high-quality commonsense knowledge graph from scratch without existing datasets, including a low-cost filtering method, (2) the comparison of prompting for humans and an LLM, and (3) the publication of the resulting Japanese commonsense knowledge graph and commonsense generation models.

\section{Related Work}

\begin{figure*}[t]
    \centering
    \begin{subfigure}[b]{0.4\linewidth}
        \centering
        \includegraphics[width=\textwidth]{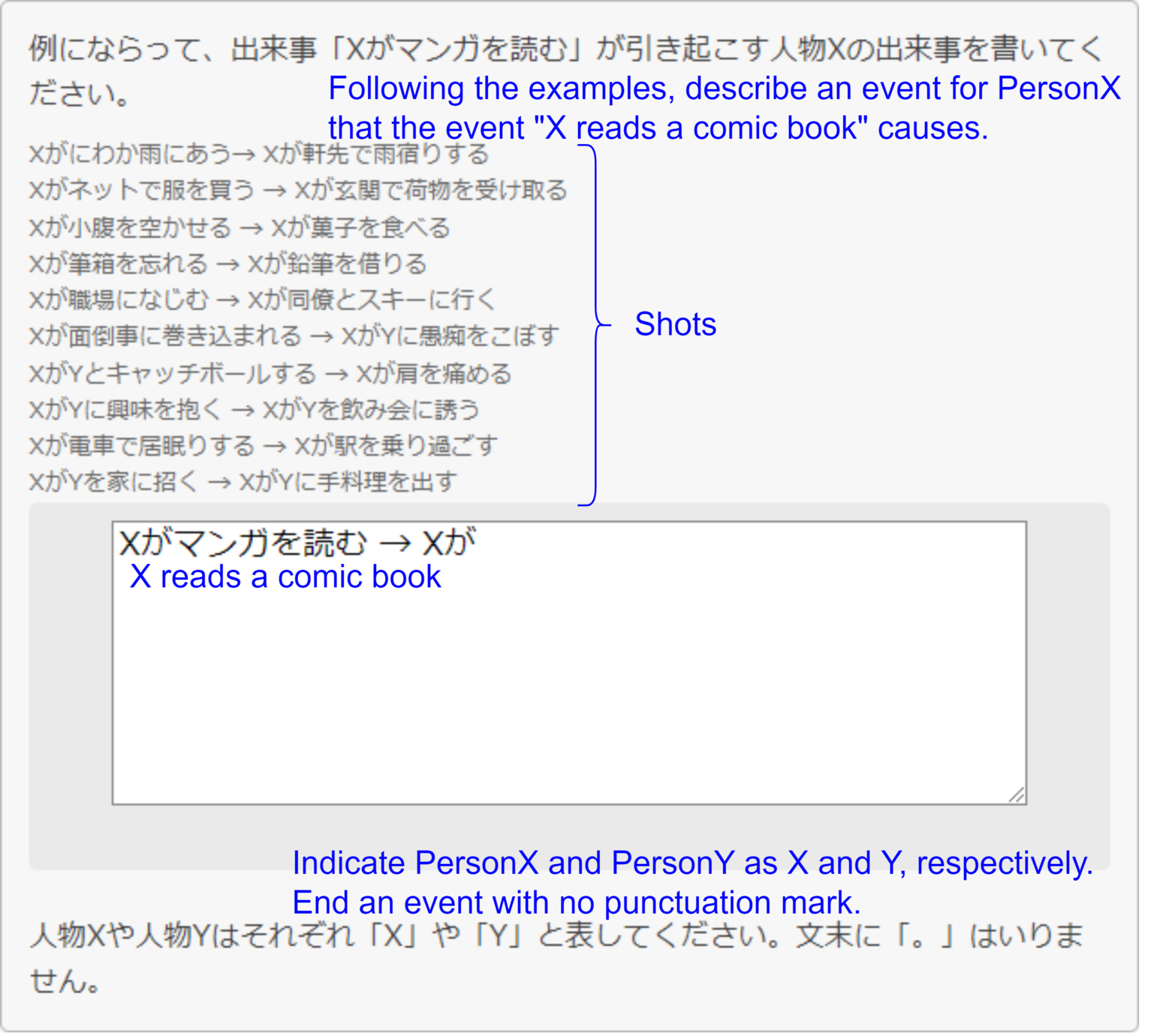}
        \caption{Crowdsourcing interface to acquire inferences (xEffect)}
        \label{fig:prompt_for_human}
    \end{subfigure}
    \hfill
    \begin{subfigure}[b]{0.4\linewidth}
        \centering
        \includegraphics[width=\textwidth]{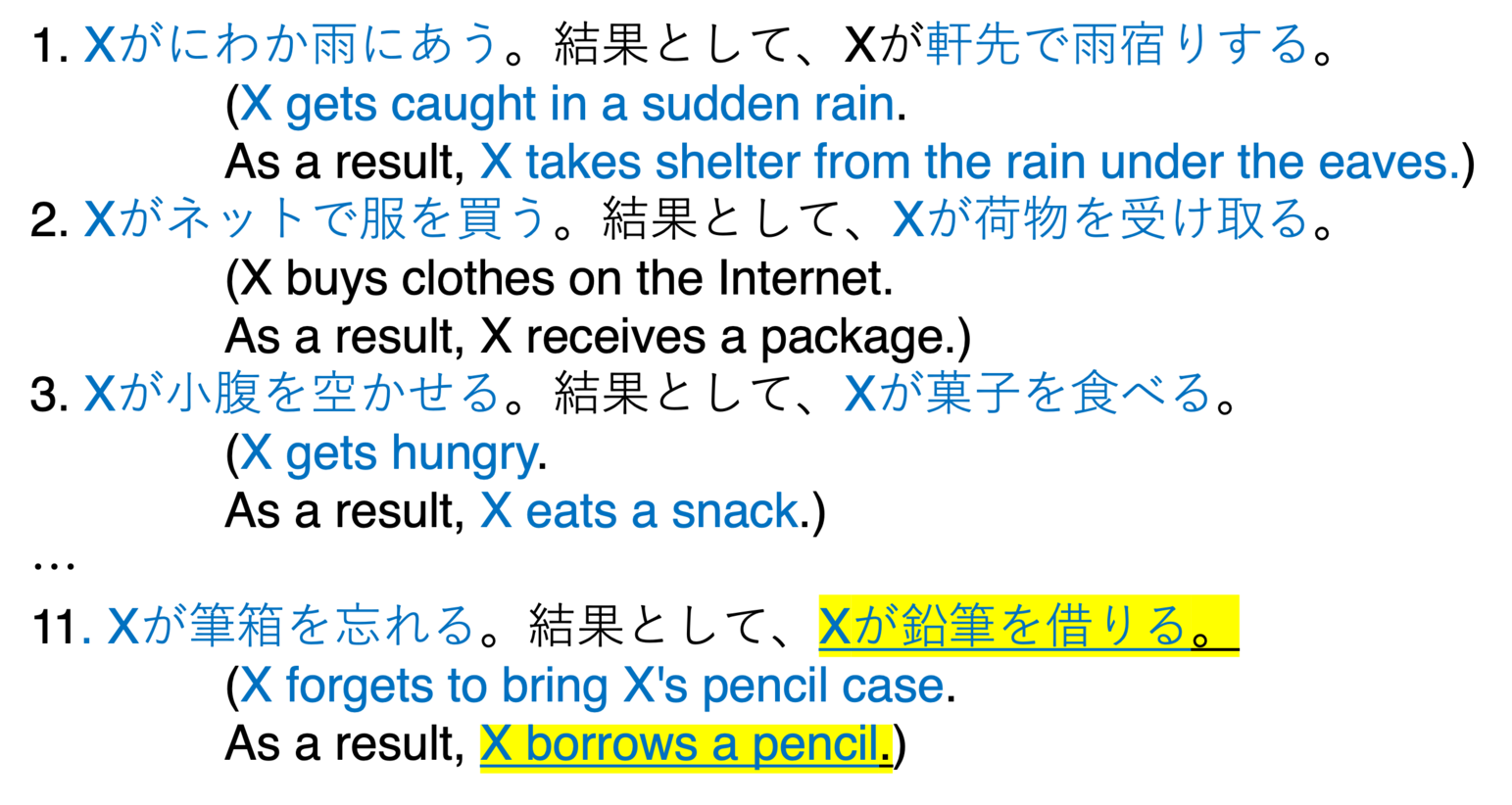}
        \caption{An example of prompts for generating inferences from HyperCLOVA. The underlined parts are generated.}
        \label{fig:prompt_for_LLM}
    \end{subfigure}
    \caption{Comparison of \textit{prompting} between for humans (Figure \ref{fig:prompt_for_human}) and for an LLM (Figure \ref{fig:prompt_for_LLM}).}
    \label{fig:comp_prompt}
\end{figure*}

\subsection{Commonsense Knowledge Datasets}

There are several knowledge bases about commonsense, from what appears in the text to what is tacit but not written in the text.
ConceptNet \citep{Speer_Chin_Havasi_2017}, for example, is a knowledge graph that connects words and phrases by relations.
GenericsKB \citep{https://doi.org/10.48550/arxiv.2005.00660} is a corpus describing knowledge of entities in natural language rather than in graph.

In some datasets, commonsense knowledge is collected in the form of question answering.
\citet{copa} acquire plausible causes and effects for premises as two-choice questions.
\citet{zellers-etal-2018-swag} provide SWAG, acquiring inferences about a situation from video captions as four-choice questions.
KUCI \citep{omura-etal-2020-method} is a dataset for commonsense inference in Japanese, which is obtained by combining automatic extraction and crowdsourcing.
\citet{talmor-etal-2019-commonsenseqa} build CommonsenseQA, which treats commonsense on ConceptNet's entities as question answering.

In contrast, we propose a method to construct a structural (not QA form) graph based on events.

\subsection{Knowledge Graphs on Events}

Regarding commonsense knowledge bases, there are several graphs that focus on events.
ATOMIC \citep{Sap_Le_Bras_Allaway_Bhagavatula_Lourie_Rashkin_Roof_Smith_Choi_2019} describes the relationship between events, mental states \citep{rashkin-etal-2018-event2mind}, and personas.
\citet{Hwang_Bhagavatula_Le_Bras_Da_Sakaguchi_Bosselut_Choi_2021} merge ATOMIC and ConceptNet, proposing ATOMIC-2020.

There are also studies for leveraging context.
GLUCOSE \citep{mostafazadeh-etal-2020-glucose} is a commonsense inference knowledge graph for short stories, built by annotating ROCStories \citep{mostafazadeh-etal-2016-corpus}.
CIDER \citep{ghosal-etal-2021-cider} and CICERO \citep{ghosal-etal-2022-cicero} are the graphs for dialogues where DailyDialog \citep{li-etal-2017-dailydialog} and other dialogue corpora are annotated with inferences.

ASER \citep{https://doi.org/10.48550/arxiv.1905.00270} is an event knowledge graph, automatically extracted from text corpora by focusing on discourse.
With ASER, TransOMCS \citep{ijcai2020p554} aims at bootstrapped knowledge graph acquisition by pattern matching and ranking.

While ConceptNet and ATOMIC are acquired by crowdsourcing, ASER and TransOMCS are automatically built.
On one hand, a large-scale graph can be built easily in an automatic way, but it is difficult to obtain knowledge not appearing in the text.
On the other hand, crowdsourcing can gather high-quality data, but it is expensive in terms of both money and time.

There is a method that uses crowdsourcing and LLMs together to build an event knowledge graph \citep{west-etal-2022-symbolic}.
Although it is possible to acquire a large-scale and high-quality graph, they assume that an initial graph, ATOMIC in this case, has already been available. As stated in Section \ref{sec:intro}, our proposal is a method to build a commonsense knowledge graph from scratch without such existing graphs.

\subsection{Commonsense Generation Models}

There have been studies on storing knowledge in a neural form rather than a symbolic form.
In particular, methods of considering neural language models as knowledge bases \citep{petroni-etal-2019-language, https://doi.org/10.48550/arxiv.2204.06031} have been developed.
\citet{bosselut-etal-2019-comet} train COMET by finetuning pretrained Transformers on ATOMIC and ConceptNet, aiming at inference on unseen events and concepts.
\citet{Gabriel_Bhagavatula_Shwartz_Le_Bras_Forbes_Choi_2021} point out that COMET ignores discourse, introducing recurrent memory for paragraph-level information.

\citet{west-etal-2022-symbolic} propose symbolic knowledge distillation where specific knowledge in a general language model is distilled into a specific language model via a symbolic one.
They expand ATOMIC using GPT-3 \citep{NEURIPS2020_1457c0d6}, filter the outputs using RoBERTa \citep{https://doi.org/10.48550/arxiv.1907.11692}, and finetune GPT-2 \citep{Radford2019LanguageMA} on the filtered ones. Training data for this filtering was created through manual annotation, which is expensive.

\section{Prompting Humans and an LLM}
\label{sec:concept}


We propose a method to build a knowledge graph for commonsense inference from scratch, with both crowdsourcing and an LLM.
In our method, we first construct a small-scale knowledge graph by crowdsourcing.
Using the small-scale graph for prompts, we then extract commonsense knowledge from an LLM.
The flow of our method is shown in Figure \ref{fig:method}.
Building a knowledge graph from scratch only by crowdsourcing is expensive in terms of both money and time.
Hence, the combination of crowdsourcing and an LLM is expected to reduce the cost.

In other words, our method consists of the following two phases:
(1) collecting a small-scale graph by crowdsourcing and
(2) generating a large-scale graph by an LLM.
While crowdsourcing elicits commonsense from people, shots are used to extract knowledge from an LLM.
At this point, these phases are intrinsically the same, being considered as \textit{prompting} (Fig \ref{fig:comp_prompt}).
In the two phases, namely, we prompt people and an LLM, respectively.

We build a commonsense inference knowledge graph in Japanese. 
We focus on an event knowledge graph such as ATOMIC \citep{Sap_Le_Bras_Allaway_Bhagavatula_Lourie_Rashkin_Roof_Smith_Choi_2019} and ASER \citep{https://doi.org/10.48550/arxiv.1905.00270}.
Handling commonsense on events and mental states would facilitate understanding of narratives and dialogues. 
We use Yahoo! Crowdsourcing in the first phase and HyperCLOVA JP \citep{kim-etal-2021-changes}, an LLM in Japanese, in the second phase. We also propose a method to filter the triplets generated by the LLM at a low cost.

\subsection{Acquisition by Crowdsourcing}
\label{subsec:build_small}

We first acquire a small-scale high-quality knowledge graph by crowdsourcing.
With Yahoo! Crowdsourcing, specifically, we ask crowdworkers to write events and inferences.
In a task, we provide them with 10 shots as a prompt for each event and inference.
Note that for inferences, the prompts differ for each relation, as mentioned later.
We obtain a graph by filtering the collected inferences syntactically and semantically.

\paragraph{Events}

We ask crowdworkers to write daily events related to at least one person (PersonX).
An example of the crowdsourcing task interface is shown in Figure \ref{fig:c_event}.
The task provides instructions and 10 examples, and each crowdworker is asked to write at least one event.
After all tasks are completed, we remove duplicate events.
As a result, 257 events were acquired from 200 crowdworkers.
We manually verified that all of the acquired events have sufficient quality.

\paragraph{Inferences}

For the events collected above, we ask crowdworkers to write inferences about what happens and how a person feels before and after the events.
In this paper, the relations for inference are based on ATOMIC.\footnote{The relations are not exactly the same as those of ATOMIC. xIntent in this paper covers xIntent and xWant in ATOMIC, and tails for our xIntent and xReact may contain not mental states but events. The reason for the difference is that English and Japanese have different linguistic characteristics, i.e., it is difficult to collect knowledge in the same structure as the original.} The following four are adopted as our target relations.
\begin{itemize}
    \setlength{\itemsep}{0.1cm}
    \item xNeed: What would have happened before
    \item xEffect: What would happen after
    \item xIntent: What PersonX would have felt before
    \item xReact: What PersonX would feel after
\end{itemize}
While xNeed and xEffect are inferences about events, xIntent and xReact are inferences about mental states.

Three crowdworkers are hired per event. Given an instruction and 10 examples, each crowdworker is asked to write one inference.
An example of the crowdsourcing task interface is shown in Figure \ref{fig:c_effect}.
We remove duplicate inferences as in the case of events and then apply syntactic filtering\footnote{KNP determines if the subject is PersonX, if the tense is present, and if the event is a single sentence.} using the Japanese syntactic parser KNP\footnote{\url{https://nlp.ist.i.kyoto-u.ac.jp/?KNP}}.

\begin{table}[t]
    \centering
    \begin{tabular}{l|r|rrr}
        \hline
         & \textbf{Inst \#} & \textbf{Val \#} & \textbf{Val \%} & \textbf{IAA} \\
        \hline
        Event & 257 & \multicolumn{1}{c}{-} & \multicolumn{1}{c}{-} & \multicolumn{1}{c}{-} \\
        \hdashline
        xNeed & 504 & 402 & 79.76 & 39.85 \\
        xEffect & 621 & 554 & 89.21 & 25.00 \\
        \hdashline
        xIntent & 603 & 519 & 86.07 & 36.11 \\
        xReact & 639 & 550 & 86.07 & 31.82 \\
        \hline
    \end{tabular}
    \caption{The statistics on events and inferences acquired by crowdsourcing. \textbf{Val \#} and \textbf{Val \%} represent the number and percentage of triplets considered valid by the crowdworkers, respectively.}
    \label{tab:stats_small}
\end{table}

\begin{table*}[t]
    \centering
    \small
    \tabcolsep 3pt
    \begin{tabular}{p{0.2\linewidth}lp{0.6\linewidth}l}
        \hline
        \textbf{Head} & \textbf{Rel} & \textbf{Tail} & \textbf{Eval} \\
        \hline
        \begin{CJK}{UTF8}{min}Xが顔を洗う\end{CJK} (X washes X's face) & xNeed & \begin{CJK}{UTF8}{min}Xが水道で水を出す\end{CJK} (X runs water from the tap) & \checkmark \\
         & & \begin{CJK}{UTF8}{min}Xが歯を磨く\end{CJK} (X brushes X's teeth) & \\
         & xEffect & \begin{CJK}{UTF8}{min}Xがタオルを準備する\end{CJK} (X prepares a towel) & \checkmark \\
         & & \begin{CJK}{UTF8}{min}Xが鏡に映った自分の顔に覚えのない傷を見つける\end{CJK} (X finds an unrecognizable scar on X's face in the mirror) & \checkmark \\
         & & \begin{CJK}{UTF8}{min}Xが歯磨きをする\end{CJK} (X brushes his teeth) & \checkmark \\
         & xIntent & \begin{CJK}{UTF8}{min}スッキリしたい\end{CJK} (Want to feel refreshed) & \checkmark \\
         & & \begin{CJK}{UTF8}{min}眠いのでしゃきっとしたい\end{CJK} (Sleepy and Want to feel refreshed) & \checkmark \\
         & xReact & \begin{CJK}{UTF8}{min}さっぱりして眠気覚ましになる\end{CJK} (Feel refreshed and shake off X's sleepiness) & \checkmark \\
         & & \begin{CJK}{UTF8}{min}きれいになる\end{CJK} (Be clean) & \checkmark \\
         & & \begin{CJK}{UTF8}{min}さっぱりした\end{CJK} (Felt refreshed) & \checkmark \\
        \hline
    \end{tabular}
    \caption{Examples of inferences acquired through crowdsourcing. Triplets with \checkmark\ in the eval column were judged to be acceptable by the evaluation in Section \ref{subsec:eval}.}
    \label{tab:eg_small}
\end{table*}

The numbers of the acquired events and inferences are shown in the Inst \# column of Table \ref{tab:stats_small}.
Examples of acquired inferences are shown in Table \ref{tab:eg_small}.

\subsection{Evaluation and Filtering of Crowdsourced Triplets}
\label{subsec:eval}

To examine the qualities of the inferences acquired by crowdsourcing, we crowdsource their evaluation.
We ask three crowdworkers whether the inferences are acceptable or not and judge their acceptability by majority voting.
The evaluation is crowdsourced independently for each relation.
The inferences judged to be unacceptable by majority voting are filtered out.

The inferences collected in Section \ref{subsec:build_small} are evaluated and filtered as above.
The statistics are listed in the middle two columns of Table \ref{tab:stats_small}.
We also calculated Fleiss's $\kappa$ \footnote{As a reference, $\kappa\in[0,0.2]$ is considered to be ``slight'' agreement, $(0.2,0.4]$ is ``fair'', and $(0.4,0.6]$ is ``moderate'' \cite{10.2307/2529310}.} as an inner-annotator agreement (IAA) in the evaluation, which is shown in the rightmost column of Table \ref{tab:stats_small}. The IAAs were ``fair'' to weak ``moderate''. This is because commonsense is subjective and depends on the individual.

There are several tendencies in the inferences filtered out, i.e., judged to be unacceptable.
In some inferences, the order is reversed, as in the triplet $\langle$PersonX sleeps twice, xEffect, PersonX thinks that they are off work today$\rangle$.
Others are not plausible, as in $\langle$PersonX surfs the Internet, xNeed, PersonX gets to the ocean$\rangle$.

\subsection{Generation from an LLM}
\label{subsec:build_large}

From a small-scale high-quality knowledge graph acquired in Sections \ref{subsec:build_small} and \ref{subsec:eval}, we generate a large-scale knowledge graph with an LLM.
We use the Koya 39B model of HyperCLOVA JP as an LLM.
Both events and inferences are generated by providing 10 shots.
The shots are randomly chosen from the small-scale graph for each generation.

\paragraph{Events}

New events are generated by HyperCLOVA JP, using the events acquired in Section \ref{subsec:eval} as shots.
An example prompt for event generation is shown in Figure \ref{fig:prompt_event}.
We generate 10,000 events, remove duplicates, and apply the same syntactic filtering as in Section \ref{subsec:build_small}.

\paragraph{Inferences}

\begin{table}[t]
    \centering
    \small
    \tabcolsep 3pt
    \begin{tabular}{l|p{0.7\linewidth}}
        \hline
        \textbf{Rel} & \textbf{Template} \\
        \hline
        xNeed & \begin{CJK}{UTF8}{min}$h$ためには、$t$必要がある。\end{CJK} (To $h$, need to $t$.) \\
        xEffect & \begin{CJK}{UTF8}{min}$h$。結果として、$t$。\end{CJK} ($h$. As a result, $t$.) \\
        \hdashline
        xIntent & \begin{CJK}{UTF8}{min}$h$のは、$t$と思ったから。\end{CJK} ($h$ because felt $t$.) \\
        xReact & \begin{CJK}{UTF8}{min}$h$と、$t$と思う。\end{CJK} ($h$ then feel $t$.) \\
        \hline
    \end{tabular}
    \caption{The templates of shots for an LLM. $h$ and $t$ stand for head and tail, respectively. When generating, $t$ is extracted.}
    \label{tab:shot_format}
\end{table}

As in event generation, the inferences acquired in Sections \ref{subsec:build_small} and \ref{subsec:eval} are used as shots.
We generate 10 inferences for each event and remove duplicate triplets.
While we simply list the shots as a prompt in event generation, different prompts are used for each relation in inference generation.
An example prompt for xEffect generation is shown in Figure \ref{fig:prompt_effect}.
Shots are given in natural language, and tails are extracted by pattern matching.
Shot templates for each relation are shown in Table \ref{tab:shot_format}.
Finally, the syntactic filtering is applied to obtain the graph.


\begin{table}[t]
    \centering
    \begin{tabular}{l|r|rr}
        \hline
         & \textbf{Inst \#} & \textbf{Val \%} & \textbf{IAA} \\
        \hline
        Event & 1,429 & \multicolumn{1}{c}{-} & \multicolumn{1}{c}{-} \\
        \hdashline
        xNeed & 44,269 & 79.95 & 31.79 \\
        xEffect & 36,920 & 81.94 & 34.11 \\
        \hdashline
        xIntent & 52,745 & 84.84 & 31.37 \\
        xReact & 60,616 & 89.20 & 19.06 \\
        \hline
    \end{tabular}
    \caption{The statistics of events and inferences generated from an LLM. Val \% and IAA are the evaluation results of randomly selected inferences.}
    \label{tab:stats_large}
\end{table}

\begin{table*}[t]
    \centering
    \small
    \tabcolsep 3pt
    \begin{tabular}{p{0.2\linewidth}lp{0.6\linewidth}}
        \hline
        \textbf{Head} & \textbf{Rel} & \textbf{Tail} \\
        \hline
        \begin{CJK}{UTF8}{min}Xがコンビニへ行く (X goes to a convenience store)\end{CJK} & xNeed & \begin{CJK}{UTF8}{min}Xが財布を持っている (X has X's wallet), Xが外出する (X goes out), Xが外出着に着替える (X changes into going-out clothes), Xが財布を持って出かける (X goes out with X's wallet), Xが外へ出る (X goes outside)\end{CJK} \\
         & xEffect & \begin{CJK}{UTF8}{min}Xが買い物をする (X goes shopping), Xが雑誌を立ち読みする (X browses through magazines), XがATMでお金をおろす (X withdraws money from ATM), Xが弁当を買う (X buys lunch), Xがアイスを買う (X buys ice cream)\end{CJK} \\
         & xIntent & \begin{CJK}{UTF8}{min}何か買いたいものがある (Want to buy something), 雑誌を買う (Buy a magazine), 飲み物を買おう (Going to buy a drink), 飲み物や食べ物を買いたい (Want to buy a drink or food), なんでもある (There is everything X wants)\end{CJK} \\
         & xReact & \begin{CJK}{UTF8}{min}何か買いたいものがある (Want to buy something), 何か買う (Buy something), 何か買おう (Going to buy something), 何か買いたくなる (Come to buy something), ついでに何か買ってしまう (Buy something incidentally)\end{CJK} \\
        \hline
    \end{tabular}
    \caption{Examples of inferences generated from an LLM. For each relation, five examples are displayed.}
    \label{tab:eg_large}
\end{table*}

\begin{figure*}[t]
    \centering
    \begin{subfigure}[b]{0.45\textwidth}
        \centering
        \includegraphics[width=\textwidth]{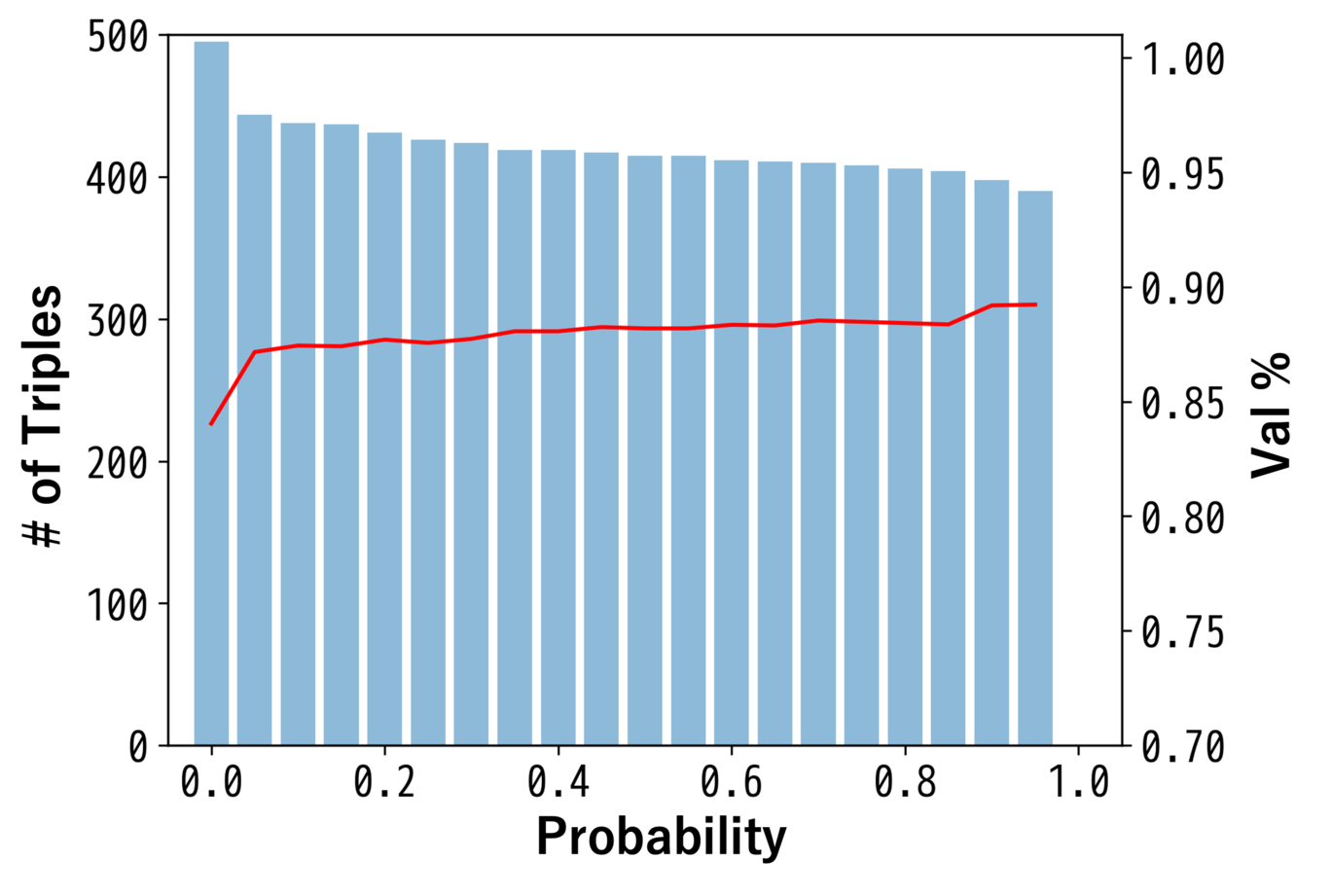}
        \caption{xNeed}
    \end{subfigure}
    \begin{subfigure}[b]{0.45\textwidth}
        \centering
        \includegraphics[width=\textwidth]{images/500_pr_11212_need.png}
        \caption{xEffect}
    \end{subfigure}
    \begin{subfigure}[b]{0.45\textwidth}
        \centering
        \includegraphics[width=\textwidth]{images/500_pr_11212_need.png}
        \caption{xIntent}
    \end{subfigure}
    \begin{subfigure}[b]{0.45\textwidth}
        \centering
        \includegraphics[width=\textwidth]{images/500_pr_11212_need.png}
        \caption{xReact}
    \end{subfigure}
    \caption{The number of inferences passed for each threshold of probability by the filter model (blue bars) and the percentage of valid inferences (red lines). }
    \label{fig:filter}
\end{figure*}

\subsection{Filtering of the Triplets Generated by LLM}
\label{subsec:filter_RoBERTa}

Because we use an LLM, some of the generated inferences are invalid. In a previous study, invalid inferences were removed by training and applying a filtering model with RoBERTa. The training data for the filtering model was made by manually annotating a portion of the large graph, which is expensive. In this section, we propose a low-cost filtering method to train filters without human annotation. This is consistent with our claim that a commonsense knowledge graph should be built for each language.

The filtering model is a binary classification model that is given a triplet and outputs whether it is valid or not. Both positive and negative examples are required for training. Since the small-scale graph used as the LLM shots was manually collected, we assume that all triplets in it are valid, and we take them as positive examples. Negative examples are not included, but we create pseudo-negative examples from the triplets in the small graph and adopt them. Details of how to create negative examples are given in Appendix \ref{sec:appendix_filter_RoBERTa}. The filtering model is constructed by finetuning Japanese RoBERTa\footnote{\url{https://huggingface.co/nlp-waseda/roberta-large-japanese}} on the training data created without human annotation. We use a RoBERTa model finetuned on an NLI task \cite{kurihara-etal-2022-jglue} in advance, following the previous study \cite{west2021symbolic}. See Appendix \ref{sec:appendix_filter_RoBERTa} for further details on the filtering model.

\subsection{The Statistics of the Large Graph}
\label{subsec:statics_large}
We obtain a large, high-quality commonsense knowledge graph from scratch by the four procedures described in the previous sections: crowdsourcing acquisition, crowdsourcing filtering, LLM scaling, and filtering with RoBERTa.
The statistics of events and inferences generated by HyperCLOVA JP are shown in Table \ref{tab:stats_large}. The results of the evaluation and the inter-annotator agreement are also shown in Table \ref{tab:stats_large}. Since the filtering threshold is a hyperparameter, we present unfiltered statistics here.
A comparison with Table \ref{tab:stats_small} indicates that the quality is as good as those written by crowdworkers.
Examples of generated inferences are shown in Table \ref{tab:eg_large}.

The generated knowledge graph in Japanese reflects the culture of Japan, such as $\langle$PersonX goes to the office, xNeed, \textit{PersonX takes a train}$\rangle$ and $\langle$PersonX eats a meal, xNeed, \textit{PersonX presses the switch on the rice cooker}$\rangle$.
This fact indicates the importance of building from scratch for a specific language, rather than translating a similar dataset in a different language, which emphasizes the value of our method proposed in this paper.

\section{Analysis on the Built Knowledge Graph}
\label{sec:build}

\subsection{Effect of Filtering for the Small Graph}

In this paper, a small-scale knowledge graph is collected as in Sections \ref{subsec:build_small} and \ref{subsec:eval}, and a large-scale knowledge graph is generated as in Section \ref{subsec:build_large}.
Here, we examine how effective the filtering for the small graph in Section \ref{subsec:eval} is.
As an experiment, we use filtered and unfiltered small-scale graphs as prompts to generate a large-scale graph.
Then, we randomly select 500 generated triplets for each relation and evaluate them by crowdsourcing as in Section \ref{subsec:eval}.
Note that the results for the filtered triplets are the same as Section \ref{subsec:build_large}.

\begin{table}[t]
    \centering
    \tabcolsep 3pt
    \begin{tabular}{l|rrrr}
        \hline
         & \textbf{xNeed} & \textbf{xEffect} & \textbf{xIntent} & \textbf{xReact} \\
        \hline
        w/o Fltr & \textbf{81.62} & 82.42 & 83.84 & 89.29 \\
        w/ Fltr & 80.81 & \textbf{85.45} & \textbf{86.06} & \textbf{90.30} \\
        \hline
    \end{tabular}
    \caption{The ratios of appropriate inferences generated by the LLM with respect to filtering.} 
    \label{tab:filter}
\end{table}

The ratios of appropriate inferences generated by the LLM with and without filtering for the small graph are shown in Table \ref{tab:filter}. We experiment with a subset of the graph.
For all relations except xNeed, filtering improves the quality of triplets.

\subsection{Effect of Filtering for the Large Graph}
To evaluate the filtering model in Section \ref{subsec:filter_RoBERTa}, 500 triplets in the large graph are manually annotated. The results of applying the filter to the test data are shown in the Figure \ref{fig:filter}. Since setting a high threshold increases the ratio of valid triplets, it can be said that we have constructed an effective filter at a low cost. 

\subsection{Comparison between Humans and an LLM}

In Section \ref{subsec:build_small}, on one hand, we asked crowdworkers to describe events and inferences.
In Section \ref{subsec:build_large}, on the other hand, we had an LLM generate them.
Here, we compare a small-scale knowledge graph by crowdsourcing and a large-scale one from an LLM, i.e., inferences generated by humans and a computer.
Because the relationships between events can be largely divided into contingent and temporal relationships \citep{bethard-etal-2008-building}, we adopt contingency and time interval as metrics for comparison.

Of the four relations, we focus on xEffect as a representative, which is a typical causal relation.
For each head of the triplets acquired by crowdsourcing in Sections \ref{subsec:build_small} and \ref{subsec:eval}, we generate three tails using the LLM in Section \ref{subsec:build_large} and compare them with the original tails.
From the 554 heads for xEffect in the small-scale graph, we obtained 586 unique inferences.

\paragraph{Contingency}

One measure is how likely a given event is to be followed by a subsequent event.
Crowdworkers are given a pair of events in an xEffect relation and asked to judge how likely the following event is to happen on a three-point scale: ``must happen,'' ``likely to happen,'' and ``does not happen.''
We ask three crowdworkers per inference and calculate the median of them.

\paragraph{Time Interval}

The other measure is the time interval between the occurrence of an event and that of a subsequent event.
As in the evaluation of contingency, crowdworkers are given a triplet on xEffect.
We ask them to judge the time interval between the two events in five levels: almost simultaneous, seconds to minutes, hours, days to months, and longer.
Finally, the median is calculated from the results of three crowdworkers.

\begin{figure*}[t]
    \centering
    \begin{subfigure}[b]{0.45\textwidth}
        \centering
        \includegraphics[width=\textwidth]{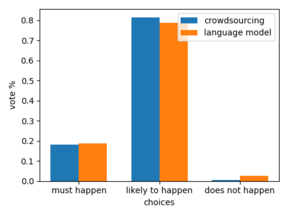}
        \caption{Contingency}
        \label{fig:comp_cont}
    \end{subfigure}
    \hfill
    \begin{subfigure}[b]{0.45\textwidth}
        \centering
        \includegraphics[width=\textwidth]{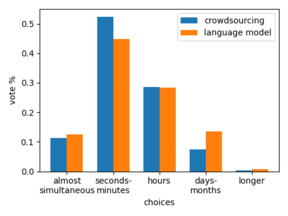}
        \caption{Time interval}
        \label{fig:comp_time}
    \end{subfigure}
    \caption{A comparison between crowdsourcing and LLM generation.}
    \label{fig:comp}
\end{figure*}

The comparison between humans and an LLM for each measure is shown in Figure \ref{fig:comp}.
Figure \ref{fig:comp_cont} shows that the subsequent events by crowdsourcing, or humans, are slightly more probable.
In Figure \ref{fig:comp_time}, the inferences generated by an LLM have a longer time interval.
This result indicates a difference in the results of prompting humans and an LLM; for xEffect, humans infer events that happen relatively soon, while an LLM infers events that happen a bit later.

\section{Japanese Commonsense Generation Models}
\label{sec:models}
To verify the quality of the graph constructed in Section \ref{sec:build} and to build a neural knowledge base, we train Japanese commonsense generation models. Japanese versions of GPT-2 \citep{Radford2019LanguageMA} and T5 \citep{JMLR:v21:20-074} are finetuned to generate inferences on unseen events. Since our preliminary experiments showed better results with GPT-2 than with T5, we report the full results with GPT-2 in this section. In preliminary experiments, the pretrained models without finetuning performed poorly, which shows that the built graph imparted commonsense reasoning ability to the models. See Appendix \ref{sec:appendix_neural_model} for the results including T5.

\subsection{Training}
Using the constructed knowledge graph, we finetune pretrained models to construct Japanese commonsense generation models. To evaluate inferences on unseen events, 150 triplets are randomly selected as the test set. For pretrained models, we adopt Japanese GPT-2\footnote{\url{https://huggingface.co/nlp-waseda/gpt2-small-japanese}} of the Hugging Face implementation \citep{wolf-etal-2020-transformers}.

As GPT-2 predicts the next word, the head and the relation are given as an input, and the model is trained to output the tail. Since the relations are not included in the vocabulary of the pretrained models, they are added as special tokens. 

\subsection{Evaluation}

\label{subsec:eval_model}
We generate inferences for the head events in the test set using the trained Japanese commonsense generation models and evaluate the inferences manually. 
Examples of the inference results are shown in Appendix \ref{sec:appendix_neural_model}.

Using crowdsourcing, we evaluate how likely the generated inferences are. We ask three crowdworkers how likely each triplet is and decide by majority vote whether each triplet is valid or not. The result is shown in Table \ref{tab:comet_eval}. Notably, the performance of GPT-2 exceeds that of HyperCLOVA. Comparing Table \ref{tab:comet_eval} with Table \ref{tab:stats_large}, which can be regarded as the performance of HyperCLOVA's commonsense generation, GPT-2 performs better for all relations. This represents the quality of the built graphs and the effectiveness of storing commonsense in the neural model.

Furthermore, as shown in Table \ref{tab:comet_eval}, the ratio of valid triplets of xNeed and xIntent are lower than xEffect and xReact, respectively. Given that xNeed is a relation that infers past events, xEffect is a future event, xIntent is a past mental state, and xReact is a future mental state, this can be attributed to the fact that we used an autoregressive model, which makes it difficult to infer in reverse order of time.

\begin{table}[t]
    \centering
    \tabcolsep 3pt
    \begin{tabular}{l|rrrr:r}
        \hline
         & \textbf{xNeed} & \textbf{xEffect} & \textbf{xIntent} & \textbf{xReact} & \textbf{Total}\\
        \hline
        \textbf{Val \%} & 84.00 & 90.00 & 90.00 & 93.33 & 89.33 \\
        \textbf{IAA} & 54.56 & 25.00 & 25.59 & 19.52 & \multicolumn{1}{c}{-} \\
        \hline
    \end{tabular}
    \caption{Evaluation scores of the commonsense generation model.}
    \label{tab:comet_eval}
\end{table}

\section{Conclusion}

We proposed a method for building a knowledge graph from scratch with both crowdsourcing and an LLM.
Based on our method, we built a knowledge graph on events and mental states in Japanese using Yahoo! Crowdsourcing and HyperCLOVA JP.
Since designing tasks for having humans describe commonsense and engineering prompts for having an LLM generate knowledge are similar to each other, we compared their characteristics.
We evaluated the graph generated by HyperCLOVA JP and found that it was similar in quality to the graph written by humans.

Furthermore, we trained a commonsense generation model for event inference based on the built knowledge graph.
We attempted inference generation for unseen events by finetuning GPT-2 in Japanese on the built graph.
The experimental results showed that these models are able to generate acceptable inferences for events and mental states.


We hope that our method for building a knowledge graph from scratch and the acquired knowledge graph lead to further studies on commonsense inference, especially in low-resource languages.

\section*{Limitations}
For acquiring a small-scale event knowledge graph and analyzing the built graph, we crowdsource several tasks using Yahoo! Crowdsourcing.
Specifically, we collect the descriptions of commonsense, filter them, and explore the characteristics of the graph by crowdsourcing.
We obtained a consent from crowdworkers on the platform of Yahoo! Crowdsourcing.

The event knowledge graph and the commonsense generation models built in this paper help computers understand commonsense.
A commonsense-aware computer, for example, can answer open-domain questions by humans, interpret human statements in detail, and converse with humans naturally.
However, such graphs and models may contain incorrect knowledge even with filtering, which leads the applications to harmful behavior.

\section*{Acknowledgements}
This work was supported by a joint research grant from LINE Corporation.

\bibliography{anthology,custom}
\bibliographystyle{acl_natbib}

\newpage
\appendix

\section{Examples of \textit{Prompting}}
\begin{figure*}[t]
    \centering
    \begin{subfigure}[b]{0.45\linewidth}
        \centering
        \includegraphics[width=\textwidth]{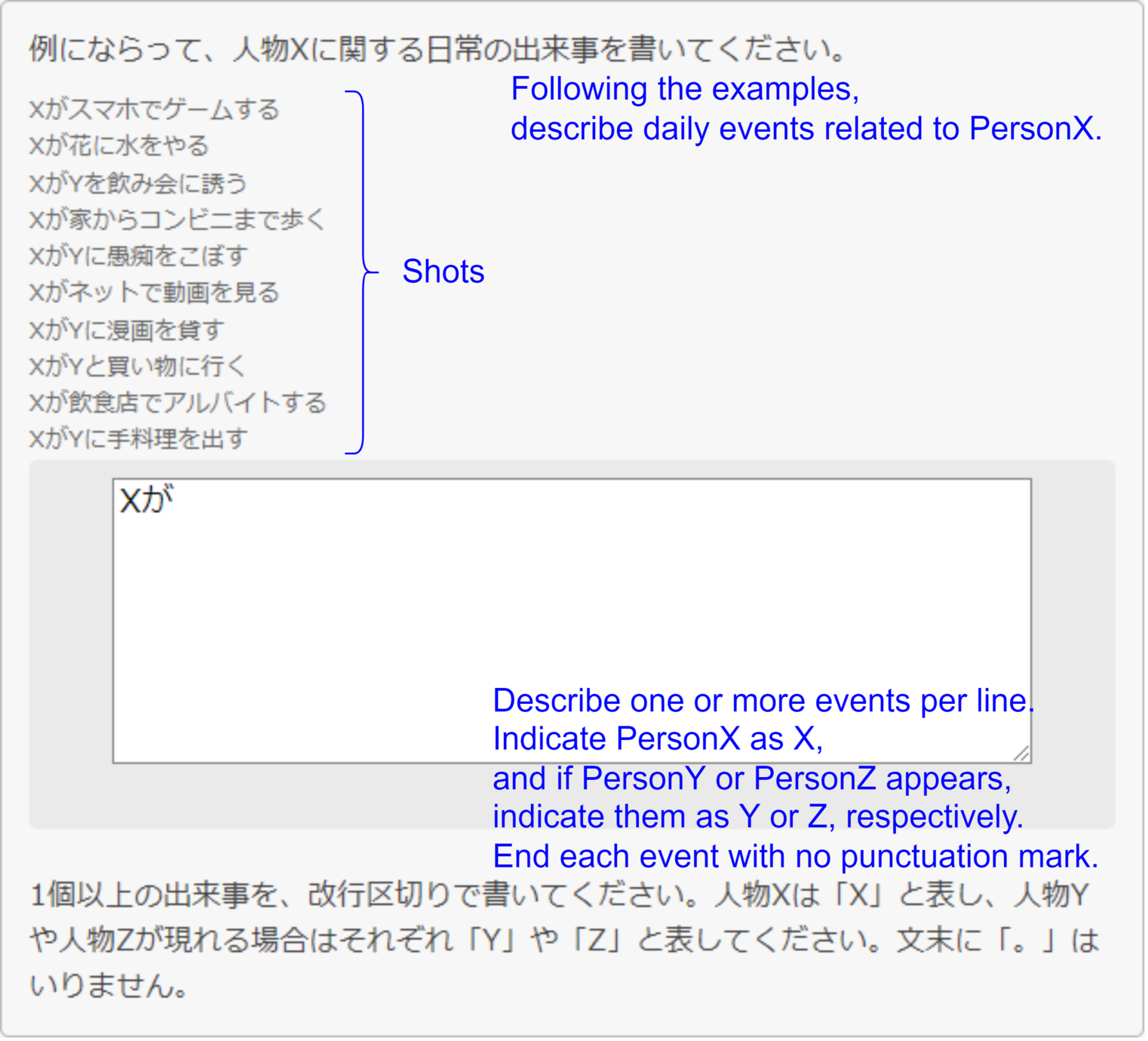}
        \caption{For events}
        \label{fig:c_event}
    \end{subfigure}
    \hfill
    \begin{subfigure}[b]{0.45\linewidth}
    \centering
    \includegraphics[width=\textwidth]{effect.png}
    \caption{For inferences (xEffect)}
    \label{fig:c_effect}
    \end{subfigure}
    \caption{Examples of crowdsourcing interfaces. Crowdworkers are asked to describe events and inferences.}
    \label{fig:c}
\end{figure*}

\begin{figure}[t]
    \centering
    \begin{subfigure}[b]{\linewidth}
        \centering
        \includegraphics[width=\linewidth]{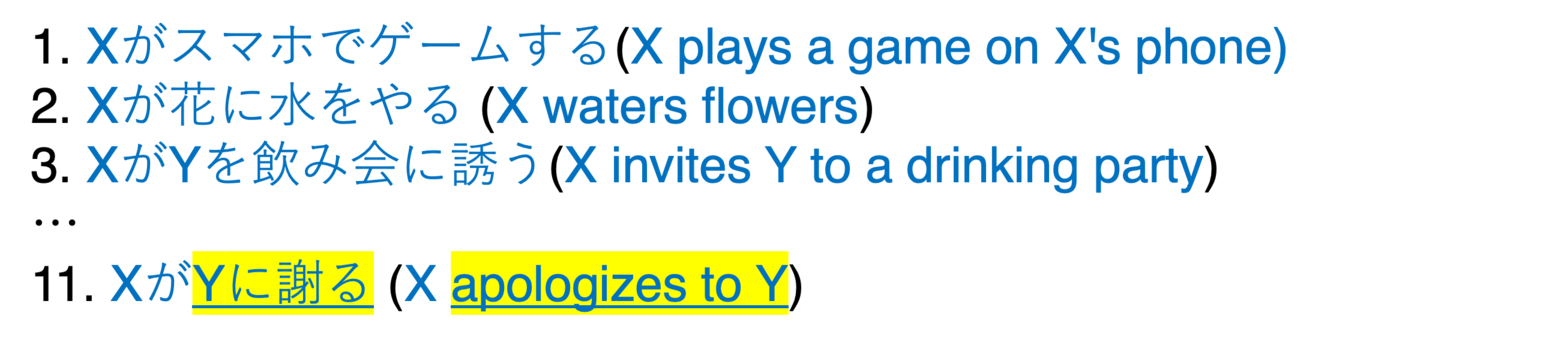}
        \caption{For events}
        \label{fig:prompt_event}
    \end{subfigure}
    \vspace{1mm} \\
    \begin{subfigure}[b]{\linewidth}
    \centering
    \includegraphics[width=\linewidth]{prompt_effect.png}
    \caption{For inferences (xEffect)}
    \label{fig:prompt_effect}
    \end{subfigure}
    \caption{Prompts for generating events and inferences from an LLM. The underlined parts are generated.}
    \label{fig:prompt}
\end{figure}

Figure \ref{fig:c} and Figure \ref{fig:prompt} are examples of \textit{prompting} both humans and an LLM, respectively.

\section{An Example of Crowdsourced Evaluation}

\begin{figure}[t]
    \centering
    \includegraphics[width=\linewidth]{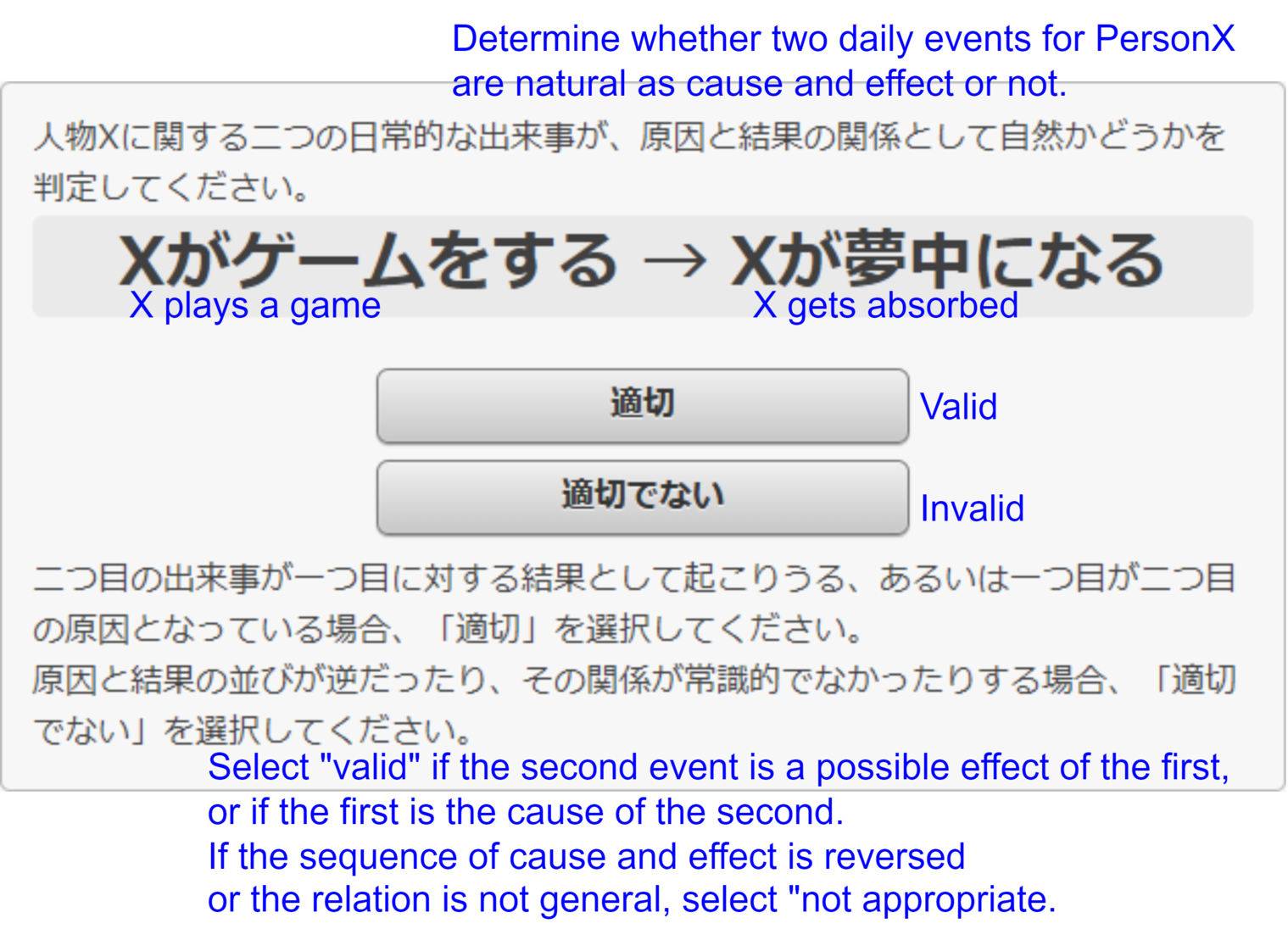}
    \caption{An example of evaluation regarding xEffect relations. We ask three crowdworkers whether a given inference is acceptable or not.}
    \label{fig:eval_effect}
\end{figure}

We evaluate and filter the inferences obtained in Sections \ref{subsec:build_small} and \ref{subsec:build_large} by crowdsourcing.
An example of the interface for evaluating an xEffect inference is shown in Figure \ref{fig:eval_effect}.

\section{Hyperparameter Details}

\begin{table}[t]
    \centering
    \begin{tabular}{l|r}
        \hline
        Max tokens & 32 \\
        Temperature & 0.5 \\
        Top-P & 0.8 \\
        Top-K & 0 \\
        Repeat penalty & 5.0 \\
        \hline
    \end{tabular}
    \caption{Hyperparameters for event and inference generation with HyperCLOVA JP.}
    \label{tab:hparam_hclova}
\end{table}

\begin{table}[t]
    \centering
    \small
    \tabcolsep 3pt
    \begin{tabular}{l|rr}
        \hline
         & \textbf{T5} & \textbf{GPT-2} \\
        \hline
        Batch size & 64 & 64 \\
        Learning rate & 5e-5 & 5e-5 \\
        Weight decay & 0.0 & 0.0 \\
        Adam betas & (0.9, 0.999) & (0.9, 0.999) \\
        Adam epsilon & 1e-8 & 1e-8 \\
        Max grad norm & 1.0 & 1.0 \\
        Num epochs & 30 & 3 \\
        LR scheduler type & Linear & Linear \\
        Warmup steps & 0 & 0 \\
        \hline
    \end{tabular}
    \caption{Hyperparameters for finetuning T5 and GPT-2 on the knowledge graph.}
    \label{tab:hparam_t5_gpt2}
\end{table}

We generate a large-scale knowledge graph using HyperCLOVA JP in Section \ref{subsec:build_large}.
The hyperparameters for the generation is shown in Table \ref{tab:hparam_hclova}.

With the built knowledge graph, we finetune Japanese T5 and GPT-2 on the task of commonsense inference in Section \ref{sec:models}.
The hyperparameters for T5 and GPT-2 are shown in Table \ref{tab:hparam_t5_gpt2}.

\section{Filtering Model Details}
\label{sec:appendix_filter_RoBERTa}
We show the details of the filtering model trained in Section \ref{subsec:filter_RoBERTa}. 

\subsection{Training Data}
First, how to make negative samples is shown. We collect 1,401 samples for xNeed, 1,750 for xEffect, 1,730 for xIntent, and 1,861 for xReact as training data. Hereafter, $G_{\mathrm{small}}$ represents the small-scale graph.
\paragraph{Negative Sample Type 1}
Pseudo-negative samples are adopted by incorrectly ordering the time series. When a valid triplet $(h,r,t) \in G_{\mathrm{small}}$ is given, an inappropriate inference can be obtained by considering $(t,r,h)$. Since the head and tail need to be swapped, it is used only for inference between event pairs.

\paragraph{Negative Sample Type 2}
By considering $(h_1,r,t_2)$ or $(h_2,r,t_1)$ from two valid triplets of the same relation $(h_1,r,t_1), (h_2,r,t_2) \in G_{\mathrm{small}}$ where $(h_1\ne h_2)$, an inappropriate inference can be obtained. Since the context is different between the head and tail, it can be an easy sample during training. For example, it is also possible to consider selecting events with similar contexts by considering the similarity between sentences to make them slightly more difficult.

\paragraph{Negative Sample Type 3}
In CSKG, there are relationships that are temporally reversed, such as xIntent and xReact. If we denote the reverse relationship for a relationship $r$ as $\mathrm{inv}(r)$, for two valid triplets $(h,r,t), (h,\mathrm{inv}(r),t') \in G_{\mathrm{small}}$ with the same head, $(h,r,t')$ can be adopted as a negative sample. Unlike Type 2, it is possible to obtain negative samples in the same context.

\subsection{Results}
The probabilities predicted by the model for each label of the test data are shown in Table \ref{tab:ave_prob}. The valid triplets yielded higher predictions, again confirming the validity of the filter model.

\begin{table}[t]
    \centering
    \begin{tabular}{c|rrr}
        \hline
        \textbf{Relation} & \textbf{valid(1)} & \textbf{invalid(2)} & \textbf{(1)-(2)} \\
        \hline
        xNeed & 0.880$\pm$0.295  & 0.626$\pm$0.450  & +0.254  \\
        xEffect & 0.367$\pm$0.447  & 0.256$\pm$0.413  & +0.111  \\
        xIntent & 0.781$\pm$0.276  & 0.643$\pm$0.300  & +0.138  \\
        xReact & 0.759$\pm$0.348  & 0.519$\pm$0.426  & +0.240  \\
        \hline
    \end{tabular}
    \caption{The average probabilities the filter model predicted for (1) the inferences deemed appropriate by crowdsourcing and (2) the inferences deemed inappropriate by crowdsourcing. Standard deviations are shown after plus/minus. The rightmost column is the difference between them.}
    \label{tab:ave_prob}
\end{table}

\section{Commonsense Generation Model Details}
\label{sec:appendix_neural_model}
\subsection{Training Details}
To confirm the appropriate model and input format, we construct some commonsense generation models with a subset of the built graph as a preliminary experiment.
For pretrained models, we adopt Japanese T5\footnote{\url{https://huggingface.co/megagonlabs/t5-base-japanese-web}} and Japanese GPT-2. 

The input format is shown in \ref{tab:input_format}. Since T5 is a seq2seq model, the head and the relation are given in the form of ``$r: h$'' as an input, and the tail is given as the correct output. The relation for T5 is changed to a natural language sentence. For example, ``xNeed'' is rewritten to ``What event occurs before this statement?'' The instructions to T5 may not be the best; prompt-engineering could improve the results. As GPT-2 predicts the next word, the head and the relation are given as an input, and the model is trained to output the tail. Since the relations are not included in the vocabulary of the pretrained models, they are added as special tokens. In the constructed knowledge graph, the subject of an event is generalized as ``X,'' but it would be better to change it into a natural expression as the input to the pretrained models. We randomly replace the subject with a personal pronoun during training and inference. To confirm this effect, we also train GPT-2 with the subject represented as ``X.'' We denote this as GPT-2$_{\mathrm{X}}$. 

\begin{table*}[t]
    \centering
    \small
    \tabcolsep 3pt
    \begin{tabular}{ll|ll}
        \hline
        \textbf{Model} & \textbf{Rel} & \textbf{Encoder Input} & \textbf{Decoder Input} \\
        \hline
        
        T5 & xNeed & \begin{CJK}{UTF8}{min}この文の前に起こるイベントは何ですか？\end{CJK}: $h$ & $t$ \\
         & & (What event occurs before this statement?: $h$) \\
         & xEffect & \begin{CJK}{UTF8}{min}このイベントの次に発生する事象は何ですか？\end{CJK}: $h$ & $t$ \\
         & & (What is the next event to occur after this event?: $h$) \\
         & xIntent & \begin{CJK}{UTF8}{min}次の文の発生した理由は何ですか？\end{CJK}: $h$ & $t$ \\
         & & (What is the reason for the occurrence of the following statement?: $h$) \\
         & xReact & \begin{CJK}{UTF8}{min}次の文の後に感じることは何ですか？\end{CJK}: $h$ & $t$ \\
         & & (What will be felt after the following statement?: $h$) \\
        \hline
        GPT-2 & xNeed & - & $h$ xNeed $t$ \\
         & xEffect & - & $h$ xEffect $t$ \\
         & xIntent & - & $h$ xIntent $t$ \\
         & xReact & - & $h$ xReact $t$ \\
        \hline
    \end{tabular}
    \caption{The input formats for training. Note that $h$ and $t$ denote a head and a tail.}
    \label{tab:input_format}
\end{table*}

\subsection{Evaluation}
\paragraph{Automatic Evaluation}
We calculate BLEU \citep{papineni-etal-2002-bleu} and BERTScore \citep{Zhang*2020BERTScore:} as automatic metrics. Table \ref{tab:score_total} shows these results. GPT-2$_{\mathrm{X}}$ and GPT-2 performed the best in BLEU and BERTScore, respectively. Table \ref{tab:ave_length_uniq_words} shows the average output length and the number of unique words for each model. The average output length of T5 is longer than those of GPT-2s, but GPT-2s have the greater numbers of unique words than T5.

\paragraph{Manual Evaluation}
Using crowdsourcing, we evaluate how likely the generated inferences are. Following the previous study \citep{west-etal-2022-symbolic}, we show crowdworkers two events (a head and a tail) and a relation. Then, we ask them to evaluate the appropriateness of the inference by choosing from the following options: ``always,'' ``often,'' ``sometimes,'' and ``never.'' The choices are displayed with an appropriate verb for each relation (e.g., ``always happens'' for xEffect). For each inference, the numbers of crowdworkers who choose ``never'' and other than ``never'' (i.e., at least ``sometimes'') are used to determine the majority vote. The acceptance rate (AR) is the proportion of inferences in which more crowdworkers choose other than ``never.'' By assigning 0 to 3 points each to ``never,'' ``sometimes,'' ``often,'' and ``always,'' we also calculate the mean point (MP) as the average score of all the inferences. Table \ref{tab:score_total} shows these results. AR is higher than 85\% for all models, indicating that the inferences for unseen events are almost correct. GPT-2 obtained the highest scores for both AR and MP.

\begin{figure}
    \centering
    \includegraphics[width=0.9\linewidth]{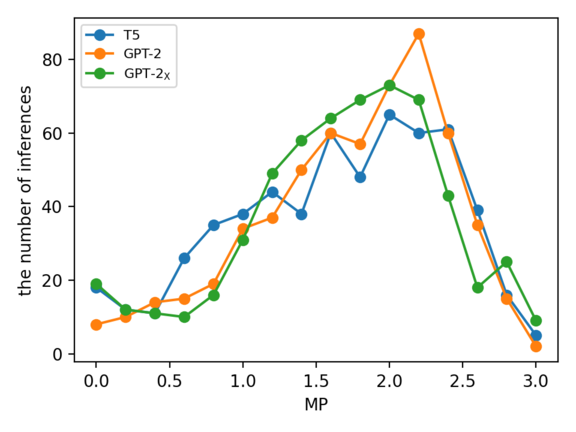}
    \vspace*{-3ex}
    \caption{The number of inferences for each MP.}
    \label{fig:num_of_inf}
\end{figure}

Although the replacement of subjects did not make a difference in AR, there is a difference in the distributions of MP as shown in Figure \ref{fig:num_of_inf}. The number of crowdworkers who chose ``never'' for the inference of GPT-2 is less than half of that for GPT-2$_{\mathrm{X}}$. This result indicates that it is better for the model to replace subjects ``X'' with personal pronouns.

Table \ref{tab:correl_auto_man} shows the correlation coefficients between the manual and automatic evaluation metrics. The correlation coefficients between the manual metrics (AR and MP) and BERTScore are positive, while those between the manual metrics and BLEU are negative or have no correlation. It seems that BERTScore, which uses vector representations, can evaluate equivalent sentences with different expressions, but BLEU, which is based on n-gram agreement, cannot correctly judge the equivalence. One of the reasons for the negative correlation in BLEU is that many inferences of the mental state consist of a single word in Japanese, such as ``tired'' and ``bored,'' for both the gold answer and the generated result. In this case, BLEU tends to be low because the words are rarely matched, but the shorter the sentences are, the easier it is for the model to generate appropriate results.

Examples of outputs are shown in Table \ref{tab:generation_example}.
We can see that the obtained outputs are acceptable to humans. The outputs vary for each model.

\begin{table}
    \centering
    \begin{tabular}{l|rrrr}
        \hline
        \textbf{Model} & \textbf{AR} & \textbf{MP} & \multicolumn{1}{c}{\textbf{BS}} & \textbf{BLEU} \\
        \hline
        T5 & 87.5 & 1.64 & 90.26 & 18.57 \\
        GPT-2 & \textbf{91.0} & \textbf{1.73} & \textbf{92.31} & 18.26 \\
        GPT-2$_{\mathrm{X}}$ & \textbf{91.0} & 1.68 & 92.03 & \textbf{18.99} \\
        \hdashline
        T5$_{\mathrm{freeze}}$ & - & - & 68.20 & 3.19 \\
        GPT-2$_{\mathrm{freeze}}$ & - & - & 69.36 & 7.33 \\
        \hline
    \end{tabular}
    \caption{Total evaluation scores. AR, MP, and BS indicate the acceptance rate, the mean point, and BERTScore, respectively. The bottom two columns show the baselines, which are not finetuned, just pretrained.}
    \label{tab:score_total}
\end{table}

\begin{table*}
    \centering
    \small
    \tabcolsep 3pt
    \begin{tabular}{l|p{0.6\linewidth}|p{0.2\linewidth}}
        \hline
        \textbf{Model} & \textbf{Input} & \textbf{Output} \\
        \hline
        T5 & \begin{CJK}{UTF8}{min}この文の前に起こるイベントは何ですか？:あなたが友人たちと旅行に出かける\end{CJK} (What event occurs before this statement?: You go on a trip with your friends) & \begin{CJK}{UTF8}{min}あなたが車を運転する\end{CJK} (You drive a car) \\
         & \begin{CJK}{UTF8}{min}このイベントの次に発生する事象は何ですか？:あなたが友人たちと旅行に出かける\end{CJK} (What is the next event to occur after this event?: You go on a trip with your friends) & \begin{CJK}{UTF8}{min}あなたが楽しい時間を過ごす\end{CJK} (You have a good time) \\
         & \begin{CJK}{UTF8}{min}次の文の発生した理由は何ですか？:あなたが友人たちと旅行に出かける\end{CJK} (What is the reason for the occurrence of the following statement?: You go on a trip with your friends) & \begin{CJK}{UTF8}{min}楽しい\end{CJK} (Have fun) \\
         & \begin{CJK}{UTF8}{min}次の文の後に感じることは何ですか？:あなたが友人たちと旅行に出かける\end{CJK} (What will be felt after the following statement?: You go on a trip with your friends) & \begin{CJK}{UTF8}{min}楽しい\end{CJK} (Have fun) \\
        \hline
        GPT-2 & \begin{CJK}{UTF8}{min}僕が友人たちと旅行に出かける\end{CJK}xNeed (I go on a trip with your friends xNeed) & \begin{CJK}{UTF8}{min}僕がパスポートを取得する\end{CJK} (I get my passport)\\
         & \begin{CJK}{UTF8}{min}僕が友人たちと旅行に出かける\end{CJK}xEffect (I go on a trip with your friends xEffect) & \begin{CJK}{UTF8}{min}僕が楽しい時間を過ごす\end{CJK} (I have a good time)\\
         & \begin{CJK}{UTF8}{min}僕が友人たちと旅行に出かける\end{CJK}xIntent (I go on a trip with your friends xIntent) & \begin{CJK}{UTF8}{min}楽しいことがしたい\end{CJK} (Want to have fun)\\
         & \begin{CJK}{UTF8}{min}僕が友人たちと旅行に出かける\end{CJK}xReact (I go on a trip with your friends xReact) & \begin{CJK}{UTF8}{min}楽しい\end{CJK} (Feel fun)\\
        \hline
    \end{tabular}
    \caption{Examples of the inferences generated by T5 and GPT-2.}
    \label{tab:generation_example}
\end{table*}

\begin{table}
    \centering
    \small
    \begin{tabular}{l|rr}
        \hline
        \textbf{Model} & \textbf{Avg Out Len} & \textbf{Uniq Word \#} \\
        \hline
        T5 & 5.29 & 392 \\
        GPT-2 & 5.03 & 451 \\
        GPT-2${}_X$ & 5.03 & 436 \\
        \hline
    \end{tabular}
    \caption{Average output length and the number of unique words.}
    \label{tab:ave_length_uniq_words}
\end{table}

\begin{table}
    \small
    \centering
    \tabcolsep 3pt
    \begin{tabular}{l|rrrr}
        \hline
         & \textbf{AR} & \textbf{MP} & \textbf{BS} & \textbf{BLEU} \\
        \hline
        AR & 1.00 & 0.75 & 0.59 & -0.11 \\
        MP & - & 1.00 & 0.43 & -0.46 \\
        BS & - & - & 1.00 & 0.30 \\
        BLEU & - & - & - & 1.00 \\
        \hline
    \end{tabular}
    \caption{Correlation coefficients between automatic and manual evaluation metrics.}
    \label{tab:correl_auto_man}
\end{table}

\end{document}